\documentclass[10pt,twocolumn,letterpaper]{article}
\usepackage[pagenumbers]{cvpr}

\ifdefined\pdfminorversion
\else
  \ifdefined\XeTeXversion
    \AtBeginDocument{}
  \fi
\fi

\usepackage[T1]{fontenc}
\usepackage{amsmath}
\usepackage{amssymb}
\usepackage{booktabs}
\usepackage{caption}
\usepackage{cuted}
\usepackage{graphicx}
\usepackage{xspace}
\usepackage{dblfloatfix}
\usepackage{float}

\newcommand{\method}{LoopCA\xspace}

\newsavebox{\resultstablebox}
\newenvironment{resultstable}[1]{%
  \begin{lrbox}{\resultstablebox}\begin{tabular}{#1}%
}{%
  \end{tabular}\end{lrbox}%
  \ifdim\wd\resultstablebox>\linewidth
    \resizebox{\linewidth}{!}{\usebox{\resultstablebox}}%
  \else
    \usebox{\resultstablebox}%
  \fi
}

\newlength{\methodsourcewidth}
\newlength{\methodsourcetmpwidth}
\newcommand{\measuremethodsource}[2]{%
  \settowidth{\methodsourcetmpwidth}{#1\quad#2}%
  \ifdim\methodsourcetmpwidth>\methodsourcewidth
    \setlength{\methodsourcewidth}{\methodsourcetmpwidth}%
  \fi
}
\newcommand{\methodsource}[2]{%
  \makebox[\methodsourcewidth][l]{#1\hfill#2}%
}

\usepackage{placeins}

\definecolor{cvprblue}{rgb}{0.21,0.49,0.74}
\usepackage[pagebackref,breaklinks,colorlinks,allcolors=cvprblue]{hyperref}
\hypersetup{
    pdftitle={LoopCA: Joint Embodied 4D World Modeling with Looped Computation and Aligned Camera-Action Control},
    pdfauthor={Shu Han, Bohan Li, Shuaiwei Wang, Gen Li, Hangyu Li, Yu Dong, Ji Wu, Hao Zhao, Xiaopeng Li}
}

\title{LoopCA: Joint Embodied 4D World Modeling with Looped Computation and Aligned Camera-Action Control}
\author{%
    Shu Han\textsuperscript{1*}\quad
    Bohan Li\textsuperscript{2,7*}\quad
    Shuaiwei Wang\textsuperscript{3,7}\quad
    Gen Li\textsuperscript{4,7}\quad
    Hangyu Li\textsuperscript{1}\\[2pt]
    Yu Dong\textsuperscript{1}\quad
    Ji Wu\textsuperscript{5\textdagger}\quad
    Hao Zhao\textsuperscript{6,7\textdagger}\quad
    Xiaopeng Li\textsuperscript{1\textdagger}\\[5pt]
    \normalsize \textsuperscript{1}University of Wisconsin--Madison\quad
    \normalsize \textsuperscript{2}Shanghai Jiao Tong University\quad
    \normalsize \textsuperscript{3}Jilin University\\[2pt]
    \normalsize \textsuperscript{4}Zhejiang University\quad
    \normalsize \textsuperscript{5}Wuhan University\quad
    \normalsize \textsuperscript{6}AIR, Tsinghua University\quad
    \normalsize \textsuperscript{7}PrimalVerse
}

\begin{document}
\twocolumn[{%
\maketitle
\vspace{-27pt}
\begin{minipage}{\textwidth}
    \centering
    \includegraphics[width=\linewidth]{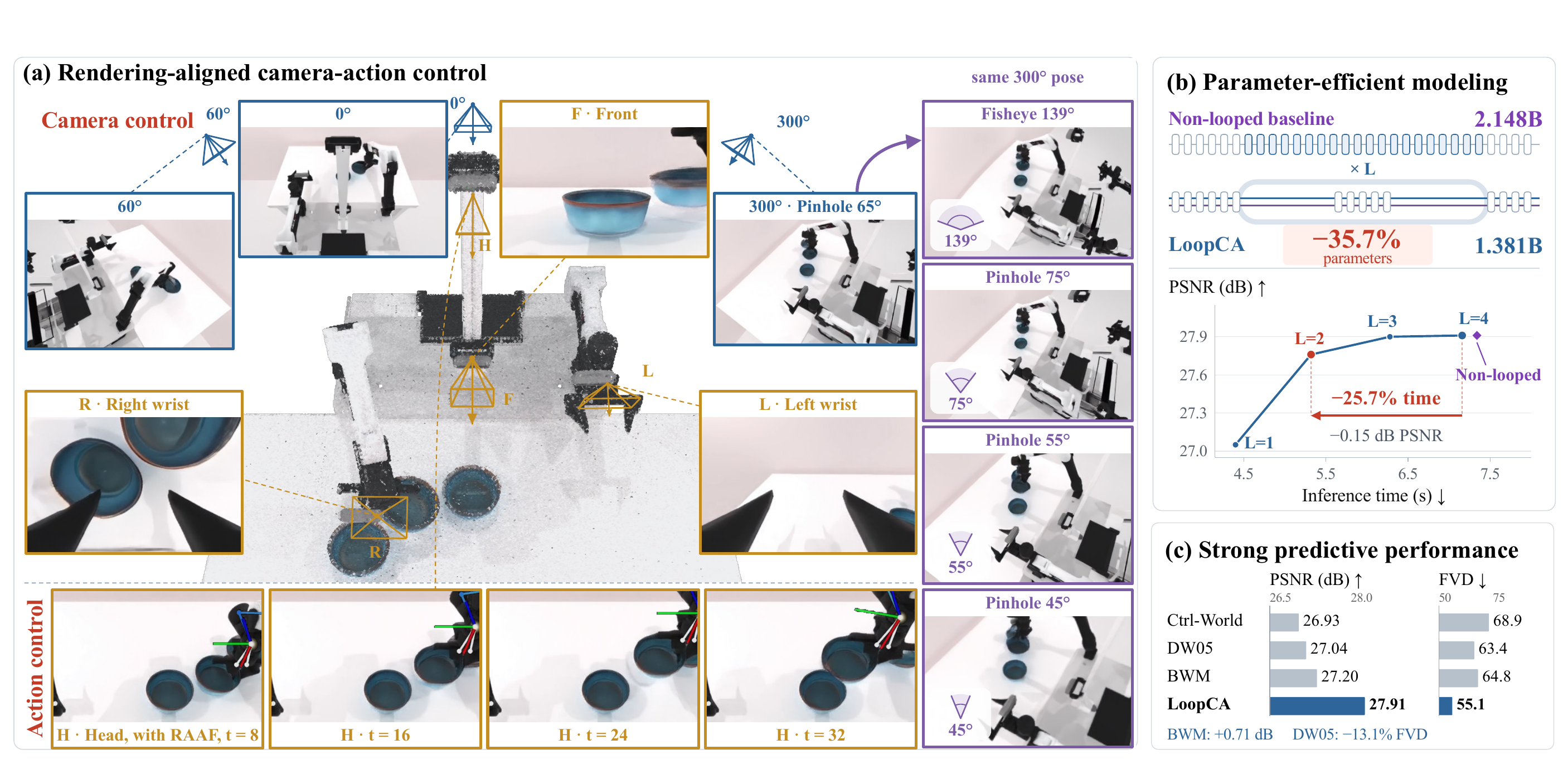}

    \captionsetup{hypcap=false}
   
\captionof{figure}{
\method\ predicts RGB videos and pointmaps under diverse camera and robot motion, from a single checkpoint with selectable inference depth.
(a) Generated views under different cameras and actions.
(b) Looping reduces model parameters by 35.7\% relative to the non-looped baseline and two-loop inference reduces latency by 25.7\% with little quality loss (see Table~\ref{tab:number_of_loops}).
(c) Results on the benchmark.
}

    \label{fig:teaser}
    \captionsetup{hypcap=true}
\end{minipage}

}]
\begingroup
\renewcommand{\thefootnote}{\fnsymbol{footnote}}
\footnotetext[1]{These authors contributed equally.\quad
\textsuperscript{\textdagger}Co-corresponding authors.}
\endgroup
\begin{abstract}

Anticipating how actions and cameras reshape the observed world is central to embodied intelligence. Joint embodied 4D world models achieve this by predicting future appearance and 3D geometry, serving as learned simulators for policy training and data generation.
Video diffusion Transformers have become the mainstream backbone for such models, but face two challenges in robotics. First, joint appearance-geometry generation increases both parameters and latency with depth, making deployment on robots costly. Second, generalized camera-action control remains challenging, as actions and cameras are heterogeneous modalities lacking unified alignment.
We present \textbf{\method}, a looped dual-stream diffusion Transformer for efficient and accurate 4D world modeling under generalized camera-action control. \method\ jointly generates RGB videos and pointmaps through dual streams that share a backbone, using stop-gradient laterals to pass RGB context to geometry. A looped core with logical-depth-specific conditioning and multi-exit supervision reduces backbone parameters and enables adjustable inference depth for lower latency. We render camera parameters and robot motion onto a shared pixel grid as per-pixel Plücker ray fields and Rendering-Aligned Action Fields (RAAFs), aligning camera and action conditions. 
On a camera-diverse RoboTwin benchmark, \method achieves \textbf{27.91 dB} PSNR and \textbf{3.67 cm} 3D point error while reducing parameters by \textbf{35.7\%} relative to the non-looped baseline. Furthermore, two-loop inference reduces latency by \textbf{25.7\%} with only a 0.15 dB PSNR drop, and augmenting policies with \method-generated views raises manipulation success by \textbf{61.8} percentage points. These results establish \method\ as an efficient, accurate, and controllable 4D world model for robotics.
\end{abstract}

\section{Introduction}
\label{sec:introduction}

Robots must anticipate how their actions and the observing camera together reshape what they see. Embodied 4D world models address this by predicting future observations under camera and action changes, serving as learned simulators~\cite{zhen2025tesseract,liu2025geometry4d} for policy learning and data generation. Appearance alone leaves scene structure implicit, so these models also predict explicit 3D geometry for spatial reasoning and manipulation~\cite{ke2024marigold,yang2024depthanythingv2}. Recent methods advance unified 4D generation and reconstruction~\cite{mi2025one4d,fang2025worldreel,chen20254dnex,zhang2024wvd} on video diffusion Transformers~\cite{wan2025wan,peebles2022dit,yang2025cogvideox}, but joint appearance-geometry prediction remains costly and hard to control under camera and action variation.

\textbf{One difficulty is the cost of joint 4D modeling.}
Joint appearance-geometry prediction typically duplicates the backbone or adds a separate branch~\cite{byungki2025jointdit,zhang2025dualcamctrl}, inflating parameters and latency as the underlying backbones keep growing.
Looped Transformers reuse backbone blocks to reduce parameter storage at a given executed depth~\cite{chng2026looped,goyal2026elt}.
Extending looped computation to joint generation means keeping per-stream states and their context exchange while camera-action guidance stays available at every reuse.
Robotic platforms also vary in available compute, so one checkpoint should trade computation for quality rather than maintaining a separate model per budget~\cite{graves2016act,banino2021pondernet,xin2020deebert,zhou2020patience}.

\textbf{A second challenge is achieving generalized joint control over camera parameters and robot motion.} Robotic cameras vary in extrinsics, intrinsics, and lens models, from pinhole to fisheye. Camera-control methods extend viewpoint guidance to diverse camera models~\cite{he2025cameractrlii,bahmani2024vd3d,bahmani2024ac3d,wang2024akira,zhang2025unifiedcamera,jin2026crepe}, while action-rendering methods align robot motion with image space using camera-projected masks, kinematic fields, or skeletons~\cite{chen2026bridgev2w,yang2026eawm,wu2026oscar,alzayer2026masked}. Building on these advances, we seek a shared camera-action interface across pinhole and fisheye optics for joint appearance-geometry prediction. Because the same 3D motion has different image-space trajectories and visibility under different cameras, control must keep kinematic and optical conditions aligned.

To address these challenges, we introduce \method, a looped dual-stream diffusion Transformer for efficient and accurate joint 4D world modeling with generalized camera-action control (\Cref{fig:teaser}).
Given an RGB reference, task text, camera parameters, and a robot trajectory, \method predicts RGB videos and pointmaps over time.
The two streams share a Transformer backbone with separate latent states, geometry adapters, and RGB-to-pointmap laterals.
A looped core with logical-depth-specific conditioning reuses blocks across streams and depth to cut parameters, with multi-exit supervision for shallower, lower-latency inference.
For unified control, per-pixel Plücker ray fields encode observation geometry under arbitrary intrinsics and extrinsics, and robot trajectories are rendered as Rendering-Aligned Action Fields (RAAFs) on the same grid to keep both signals geometrically aligned.

On a camera-diverse RoboTwin benchmark, \method\ achieves 27.91 dB PSNR, 3.67 cm 3D point error, while using 35.7\% fewer parameters than the non-looped baseline. Selecting two instead of four loops cuts latency by 25.7\% with only a 0.15 dB drop in PSNR. Moreover, augmenting downstream policies with \method-synthesized views boosts manipulation success by 61.8 percentage points. These results show that \method\ achieves strong performance in joint appearance-geometry prediction under generalized camera-action control with an adjustable compute budget.

Our main contributions are: 
(i) \textbf{Coupled Dual-Stream 4D Generation.} We formulate joint appearance-geometry world modeling in a shared video latent space, maintaining modality-specific states with geometry adapters and stop-gradient RGB-to-pointmap laterals for coherent RGB and pointmap synthesis.
(ii) \textbf{Looped Multi-Exit Computation.} We introduce a looped architecture with cross-stream and cross-depth block reuse, logical-depth-specific conditioning, and multi-exit supervision, reducing backbone parameters and enabling adjustable inference depth.
(iii) \textbf{Rendering-Aligned Camera-Action Control.} We derive per-pixel Plücker ray fields from arbitrary camera parameters and render robot trajectories into RAAFs on the same pixel grid, eliminating geometric misalignment across diverse cameras and robot actions.
\begin{figure*}[t]
    \centering
    \includegraphics[width=0.90\textwidth]{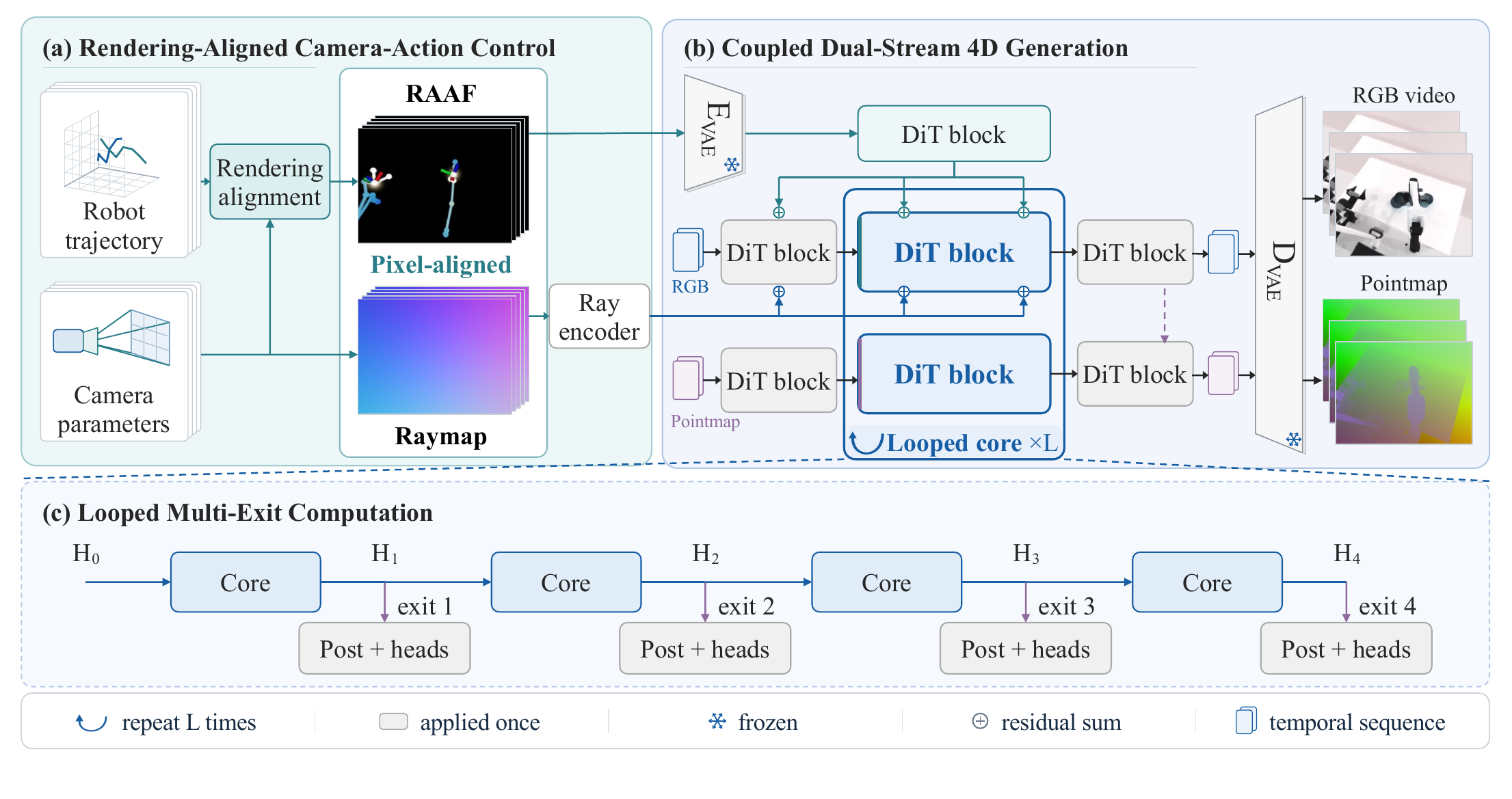}
    \caption{\textbf{Overview of \method.}
    (a) Rendering-aligned camera-action control: per-pixel Pl\"ucker ray fields
    represent camera geometry, and RAAFs are rendered on the same image grid.
    (b) Coupled dual-stream generation: a DiT with separate RGB and pointmap
    states, shared backbone blocks, and stop-gradient laterals.
    (c) Looped multi-exit computation: the shared core repeats $L$ times,
    each repetition yielding a supervised exit.}
    \label{fig:overview}
\end{figure*}

\section{Related Work}
\label{sec:related_work}

\paragraph{4D world models.}
Video world models enable scene generation~\cite{du2023universalpolicies,wu2023videopretraining,li2024uniscene,chen2024diffusionforcing,li2024occscene,nvidia2025cosmos,wan2025wan,ho2022videodiffusion,blattmann2023videoldm,blattmann2023stablevideo,bruce2024genie}, robotic simulation~\cite{ha2018worldmodels}, and policy learning~\cite{hafner2019planet,hafner2020dreamer,hafner2021dreamerv2,hafner2023dreamerv3} with actions and language~\cite{yang2023unisim,xiang2024pandora,zhu2024irasim,jang2025dreamgen,shang2026worldarena}.
4D models expose scene structure by predicting RGB with depth, normals, or pointmaps for spatial consistency, novel views, and 4D generation~\cite{zhen2025tesseract,liu2025geometry4d,zhang2024wvd,mi2025one4d,chen20254dnex,fang2025worldreel,li2025omninwm,guo2025dist4d,li2024uniscene,li2025scalingoccupancy,li2024occscene,zou2025mudg,pumarola2021dnerf,park2021hypernerf,wu20244dgs}.
Pointmaps and 3D features also support reconstruction, correspondence, geometric world modeling, ray-based pose estimation, and robot control~\cite{wang2023dust3r,leroy2024mast3r,zhang2024monst3r,wang2025vggt,wang2025cut3r,jiang2025geo4d,wu2025geometryforcing,zhang2024camerasasrays,tong2026pointaction}.
One4D uses modality-specific LoRA adapters and lightweight links for coupled RGB-pointmap modeling~\cite{mi2025one4d}. Our focus is looped parameter sharing and generalized camera-action control for joint RGB-pointmap prediction.

\paragraph{Generalized camera-action conditioning.}
Camera-controlled video generation provides explicit viewpoint guidance, but pose alone leaves intrinsics and lens-dependent observation geometry unspecified.
MotionCtrl separates camera and object motion~\cite{wang2023motionctrl}, CameraCtrl uses Pl\"ucker rays~\cite{he2024cameractrl}, and CamI2V employs epipolar attention~\cite{zheng2024cami2v}.
Later methods explore single-/multi-view camera control~\cite{he2025cameractrlii,bahmani2024vd3d,bahmani2024ac3d,xu2024camco,xu2024cavia}, geometry-guided view synthesis and rerendering~\cite{ren2025gen3c,yu2024viewcrafter,bai2025recammaster,liu2025lightx,liu2023zero123,voleti2024sv3d,wu2024reconfusion,gao2024cat3d}, and broader camera models~\cite{wang2024akira,zhang2025unifiedcamera,jin2026crepe}.
Rendered action conditions also guide interaction generation, from embodiment masks~\cite{chen2026bridgev2w}, kinematic guidance~\cite{yang2026eawm}, and skeletons~\cite{wu2026oscar} to pixel-space trajectories~\cite{alzayer2026masked}, and extend to surgical and human-object tasks~\cite{li2026surgical,gao2026pam}.
BridgeV2W and EA-WM explicitly project actions under camera calibration~\cite{chen2026bridgev2w,yang2026eawm}. Building on camera-projected action guidance, \method\ combines per-pixel Pl\"ucker rays and trajectory-rendered RAAFs across pinhole and fisheye cameras within a looped joint RGB-pointmap model.

\paragraph{Looped Transformers.}
Looped Transformers improve parameter efficiency by sharing blocks across depth~\cite{dehghani2018universal,lan2019albert,geiping2025latent,bai2019deq,bae2024relaxed}; language-model studies also examine loop placement and specialization~\cite{ekbote2026loop}. Visual generators study elastic looping, looped flow, and early exits~\cite{goyal2026elt,derakhshani2026lift,moon2024earlyexiting}.
Looped-DiT combines looped blocks, intermediate supervision, and self-modulating attention~\cite{chng2026looped}.
Building on these looped generators, \method\ extends looped computation to coupled RGB-pointmap video generation, with separate modality states, geometry adapters, and camera-action conditioning at each logical depth.
\section{Method}
\label{sec:method}

As shown in Figure~\ref{fig:overview}, \method\ is a looped dual-stream
diffusion world model. Given an RGB reference, task text, per-frame camera
parameters, and a supplied robot trajectory, it jointly predicts a future
RGB video and the pointmap of visible surfaces. The RGB
and pointmap streams share backbone blocks while keeping separate latent
states and geometry adapters; rendering-aligned camera-action conditioning
renders the supplied trajectory into action fields on the same pixel grid
as the camera rays, and a looped core with multi-exit supervision supports
selectable inference depth.

\subsection{Problem Definition}
\label{sec:problem_formulation}

Given an RGB reference $I_0$, task text $c$, camera parameters
$\mathcal C_{0:T}$, and supplied robot trajectory $Q_{0:T}$, we model
\begin{equation}
    p_\theta\!\left(V_{0:T},P_{0:T}
    \mid I_0,c,\mathcal C_{0:T},Q_{0:T}\right),
    \label{eq:problem_formulation}
\end{equation}
where $V$ is an RGB video and $P_t(u)\in\mathbb R^3$ stores each valid
pixel's world-coordinate position. Pointmaps describe visible surfaces over
time without assuming complete coverage or persistent point identities.
The sequence reconstructs the reference at $t=0$ and predicts future frames
$t=1,\ldots,T$. At inference, the model receives a reference RGB image,
task text, the camera parameter sequence $\mathcal C_{0:T}$, and the robot
trajectory $Q_{0:T}$, both including the initial state at $t=0$.

\subsection{Rendering-Aligned Camera-Action Control}
\label{sec:camera_control}
\label{sec:raaf}

Camera and action control are coupled in a robot world model, where a
trajectory lives in 3D kinematic coordinates, but its image-space appearance
depends on the observing camera, so the same motion produces different
trajectories across different camera parameters. We therefore render the camera
rays and the action fields under the same target camera onto the same pixel grid, keeping the two conditions aligned as the camera changes.

\paragraph{Camera rays.}
We represent the observation geometry through a per-pixel Pl\"ucker ray
field. Inverse projection gives each pixel's unit direction
$d_t^{\mathrm{cam}}(u)$, using the intrinsic matrix $K_t$ for pinhole cameras
and per-pixel ray directions for fisheye cameras. For camera-to-world
transform $T_t$, we use
\begin{equation}
    \bar T_t=T_{\mathrm{head}}^{-1}T_t=[\bar R_t\mid\bar o_t],
    \label{eq:camera_frame}
\end{equation}
where $T_{\mathrm{head}}$ is the fixed head-camera pose shared across
episodes, preserving sensor-placement differences without per-clip pose
normalization. The Pl\"ucker ray field is
\begin{equation}
\begin{aligned}
    \bar d_t(u)&=\bar R_t d_t^{\mathrm{cam}}(u),\\
    \mathcal R_t(u)&=
    \bigl(\bar o_t\times\bar d_t(u),\,\bar d_t(u)\bigr).
\end{aligned}
    \label{eq:plucker_ray}
\end{equation}
This six-channel interface supports fixed and moving cameras; pointmap
targets remain in metric world coordinates. Intrinsic edits change pixel
directions; extrinsic edits change ray origins and orientations.

\paragraph{Action fields.}
We represent the robot motion by rendering the trajectory into
Rendering-Aligned Action Fields (RAAFs) on the same pixel grid as the rays.
At each frame, the trajectory $Q_t=(\mathbf q_t, g_t, \boldsymbol\xi_t)$ provides the joint configuration $\mathbf q_t$, the gripper state $g_t$, and the end-effector pose $\boldsymbol\xi_t$; because the pretrained video DiT consumes visual latents rather than kinematic states, we render these quantities as an image. Forward kinematics
first maps $\mathbf q_t$ to the world-frame pose $\mathbf X^W_k(t)$ of every
link $k$,
\begin{equation}
    \mathbf X^W_k(t)=\mathrm{FK}_k(\mathbf q_t),
    \label{eq:raaf_fk}
\end{equation}
whose positions form the keypoint set $\mathcal J_t=\{\mathbf p^W_k(t)\}$
marking the arm and gripper. We then project each world-frame keypoint
$\mathbf p^W\in\mathcal J_t$ to its image position $u_t$ via the camera projection operator $\Pi(\cdot)$ parameterized by the camera-to-world transform $T_t$ and camera intrinsics $\mathcal K_t$ (the pinhole matrix $K_t$ or fisheye per-pixel ray directions),
\begin{equation}
    u_t=\Pi(\mathbf p^W;T_t,\mathcal K_t),
    \label{eq:raaf_projection}
\end{equation}
and rasterize the projected keypoints together with $g_t$ and $\boldsymbol\xi_t$ into the RGB action field $A_t$ using color-coded geometric primitives encoding link connectivity, end-effector orientation, and gripper state:
\begin{equation}
    A_t=\operatorname{Render}\bigl(\mathcal J_t,g_t,\boldsymbol\xi_t;T_t,\mathcal K_t\bigr).
    \label{eq:raaf_rendering}
\end{equation}
RAAF, RGB, and pointmaps share temporal sampling and an image grid. Camera
edits update both rays and projected motion, maintaining their rendering
alignment.

\subsection{Dual-Stream Backbone}
\label{sec:architecture}

\paragraph{Dual-stream architecture.}
We build on a pretrained video diffusion Transformer
(DiT)~\cite{peebles2022dit}. RGB and pointmaps are encoded and decoded by
the same frozen video variational autoencoder (VAE), following latent
diffusion~\cite{rombach2021latentdiffusion}, with world coordinates
normalized to $[-1,1]$ under fixed bounds before encoding. The two
modalities form two separate 16-channel latent streams that share the
backbone blocks but keep their own patch embeddings and velocity heads;
geometry residual adapters specialize the shared blocks for geometry,
reusing the pretrained latent interface without a separately trained
geometry codec.

\paragraph{Cross-modal coupling.}
At logical layer $l$, a learned lateral $\phi_l$ transfers visual context
from the RGB hidden state $H_{l,\mathrm{rgb}}$ to the pointmap hidden state
$H_{l,\mathrm{pm}}$:
\begin{equation}
    H_{l,\mathrm{pm}}\leftarrow H_{l,\mathrm{pm}}
    +\phi_l\!\left(\operatorname{sg}(H_{l,\mathrm{rgb}})\right),
    \label{eq:rgb_pointmap_lateral}
\end{equation}
Here, $\operatorname{sg}$ denotes stop-gradient: geometry losses update shared weights through the pointmap stream without backpropagating through the laterals into RGB features. Geometry thus borrows visual context while the RGB stream is shielded from direct gradient interference.

\paragraph{Condition injection.}
Following prior work on camera conditioning~\cite{he2024cameractrl}, a ray
encoder maps the Pl\"ucker ray field to DiT-aligned token features. In
parallel, we condition on the RAAF through a video-conditioning
branch~\cite{jiang2025vace} that processes the noisy RGB tokens together
with the VAE-encoded RAAF and produces residual features. Zero-initialized projections inject the camera and action residuals into the RGB stream at uniformly spaced logical layers across the backbone, and the laterals pass this context to geometry.
The video-conditioning blocks inherit the
corresponding pretrained DiT weights, so predictions are preserved when the branch is introduced.

\subsection{Looped Computation and Conditioning}
\label{sec:looped_dit}

We adapt looped computation~\cite{chng2026looped} by sharing backbone
blocks across both streams and loop depth while keeping separate
states and depth-specific geometry and conditioning modules.

\paragraph{Looped backbone.}
The pretrained backbone is split into a fixed pre stack $F_{\mathrm{pre}}$, a
shared core $F_{\mathrm{shared}}$, and a post stack $F_{\mathrm{post}}$. For
each modality $m\in\{\mathrm{rgb},\mathrm{pm}\}$ with hidden state $H_m^{(k)}$
at fixed diffusion time $s$,
\begin{equation}
\begin{aligned}
    H_m^{(0)} &= F_{\mathrm{pre}}(H_m^{\mathrm{in}};s),\\
    H_m^{(k)} &= F_{\mathrm{shared}}(H_m^{(k-1)};s,\Gamma_k),\\
    \widehat v_m^{(k)} &=
    G_m\!\left(F_{\mathrm{post}}(H_m^{(k)};s)\right),
\end{aligned}
    \label{eq:looped_recurrence}
\end{equation}
where $H_m^{\mathrm{in}}$ is the input, $\widehat v_m^{(k)}$ the velocity predicted by the modality-specific heads $G_m$, and $\Gamma_k$ the logical-depth-specific conditioning (comprising camera-action residual injections and timestep embeddings) in loop $k$, yielding a total unrolled logical depth of $N_{\mathrm{pre}} + L \cdot N_{\mathrm{shared}} + N_{\mathrm{post}}$; text conditioning is omitted for brevity.
Core states propagate without detachment, and shared post/head readouts do
not feed back into the loop. Only the core uses exclusive self-attention
(XSA)~\cite{zhai2026xsa}, which removes from each token's
attention output the component along its own value direction; this
parameter-free projection is applied only within the shared core.

\paragraph{Logical-depth-specific control.}
Each core application uses its own logical-depth-specific conditioning
modules, so camera-action guidance stays available throughout the repeated
computation. Selecting the loop count $L$ runs fewer or more core
applications with the same stored weights, and every exit reads from the
shared post stack. Parameter storage therefore stays fixed while the
executed depth grows with $L$. Looping operates within each denoising step,
independently of video and diffusion time, and each supported exit receives
direct supervision.

\subsection{Joint Training}
\label{sec:training_inference}

\paragraph{Joint latent flow matching.}
For modality $m\in\{\mathrm{rgb},\mathrm{pm}\}$ with clean latent $z_{0,m}$,
flow matching~\cite{lipman2022flowmatching} builds the noisy latent $z_{s,m}$
at noise level $\sigma_s$ from independent Gaussian noise $\varepsilon_m$ and
a shared diffusion time $s$, with velocity target $v_m$:
\begin{equation}
    z_{s,m}=(1-\sigma_s)z_{0,m}+\sigma_s\varepsilon_m,
    \qquad v_m=\varepsilon_m-z_{0,m}.
    \label{eq:flow_matching_path}
\end{equation}
Exit $k$ predicts $\widehat v_m^{(k)}$ with per-modality loss $\ell_m^{(k)}$
\begin{equation}
    \ell_m^{(k)}=
    \frac{\sum_j M_{j,m}
    (\widehat v_{j,m}^{(k)}-v_{j,m})^2}
    {\max(1,\sum_j M_{j,m})},
    \label{eq:modality_loss}
\end{equation}
Index $j$ spans latent scalar entries. $M_{\mathrm{rgb}}$ is all ones;
$M_{\mathrm{pm}}$ is the validity mask downsampled by nearest-neighbor
sampling to the latent grid and broadcast across channels. With a time-dependent scheduler
weight $w(s)$, the exit loss is
\begin{equation}
    \mathcal L_k=\mathbb E\!\left[
    w(s)\bigl(\ell_{\mathrm{rgb}}^{(k)}
    +\ell_{\mathrm{pm}}^{(k)}\bigr)\right],
    \label{eq:joint_loss}
\end{equation}
Both losses supervise latent velocities.

\paragraph{Multi-exit supervision.}
For $L>1$, we train every supported depth directly by minimizing the multi-exit loss
\begin{equation}
    \mathcal L_{\mathrm{multi}}=\mathcal L_L+
    \frac{1}{L-1}\sum_{k=1}^{L-1}\mathcal L_k.
    \label{eq:loop_supervision}
\end{equation}
All exits share the example, noise, diffusion time, and targets, so the
shallow exits receive clear learning signals without teacher
distillation. Averaging the shallow losses while keeping full weight on the
deepest exit makes the deepest exit the primary objective and trains the
lower-budget predictions. Extra post-stack readouts add training
computation only; inference evaluates one exit, and single-loop ($L{=}1$) inference uses only the first exit. Training costs are detailed
in the supplement.

\section{Experiments}
\label{sec:experiments}
\raggedbottom

We evaluate \method\ on joint prediction, camera-action control, efficiency,
and downstream policy learning, and ablate its key design choices.
\subsection{Experimental Setup}
\label{sec:locam_setup}

\paragraph{Data.}
We collect 2,418 RoboTwin~\cite{mu2025robotwin,chen2025robotwin2} episodes on 50 tasks,
each rendered with two pinhole and two equidistant-fisheye cameras,
including moving views. We use 2,318 training, 50 validation, and 50 test
episodes. All views and overlapping windows of an episode stay in one
partition. Each window contains 33 frames at $240\times320$ resolution.
The Test Set has 400 windows from 50 held-out episodes across 50 tasks.
The Pose Set consists of the 123 Test Set windows with moving cameras
and is used to evaluate camera control.

\paragraph{Training details.}
We build on Wan2.1-1.3B~\cite{wan2025wan}, keeping 6
pre blocks, a 5-block shared core, and 4 post blocks from its 30-block
backbone. The core is repeated $L=4$ times. Pointmap world coordinates are normalized to
$[-1,1]$ with fixed bounds $[-2,2]\times[-1.2,0.6]\times[-0.05,2.0]$\,m.
We train in two stages. First, we train the looped backbone, camera encoder, and pointmap
modules for three epochs. Then we add a video-conditioning
branch for RAAF and continue for four epochs from the first-stage EMA. AdamW
uses learning rate $10^{-5}$ ($5\times10^{-5}$ for new modules),
500 warmup updates, effective batch 16, and EMA decay 0.999. Text, CLIP,
and VAE weights are frozen and text dropout is 0.1.
Inference uses 50 denoising steps, classifier-free guidance (CFG) scale 1,
and seed 42.

\paragraph{Compared methods.}
We compare against camera-controlled, action-conditioned, and geometry-predicting baselines, as listed in Section~\ref{sec:external_comparison}. All compared methods are fine-tuned on our training data and evaluated on the same Test Set.

\paragraph{Metrics.}
We report visual quality with PSNR (in dB), SSIM~\cite{wang2004ssim},
LPIPS~\cite{zhang2018lpips}, FID~\cite{heusel2017fid}, and
FVD~\cite{unterthiner2018fvd}. For camera accuracy, we localize generated
RGB against static ground-truth scene maps and report rotation error
(RotErr, in degrees) and translation error (TransErr, in cm). For
geometry quality, we report mean absolute 3D point error (AbsErr, in cm),
computed as the Euclidean distance between corresponding predicted and
ground-truth world-coordinate points, and the percentage of inliers
($\delta_p$, within 5\,cm).
For action control, we report WorldArena~\cite{shang2026worldarena}
Trajectory Accuracy (TrajAcc), frame-aligned end-effector position error
(EEFErr, in pixels), and normalized PCK-AUC~\cite{lee2020dream} (in \%)
over 0--20 pixels. 

\subsection{Benchmark Comparison}
\label{sec:external_comparison}
\label{sec:joint_results}
\label{sec:camera_results}
\label{sec:efficiency}

\begin{table}[!t]
\centering
\footnotesize
\setlength{\tabcolsep}{2pt}
\caption{\textbf{RGB prediction on Test Set.} C and A denote camera and action conditioning, respectively.}
\label{tab:external_comparison}
\setlength{\methodsourcewidth}{0pt}
\measuremethodsource{RealCam-I2V}{\cite{li2025realcam}~{\normalfont\scriptsize\mbox{ICCV'25}}}
\measuremethodsource{MotionCtrl}{\cite{wang2023motionctrl}~{\normalfont\scriptsize\mbox{SIGGRAPH'24}}}
\measuremethodsource{CameraCtrl}{\cite{he2024cameractrl}~{\normalfont\scriptsize\mbox{ICLR'25}}}
\measuremethodsource{TriMotion}{\cite{shin2026trimotion}~{\normalfont\scriptsize\mbox{ECCV'26}}}
\measuremethodsource{DualCamCtrl}{\cite{zhang2025dualcamctrl}~{\normalfont\scriptsize\mbox{ECCV'26}}}
\measuremethodsource{PPMem}{\cite{he2026ppmem}~{\normalfont\scriptsize\mbox{NeurIPS'26}}}
\measuremethodsource{IRASim}{\cite{zhu2024irasim}~{\normalfont\scriptsize\mbox{ICCV'25}}}
\measuremethodsource{Ctrl-World}{\cite{guo2026ctrl}~{\normalfont\scriptsize\mbox{ICLR'26}}}
\measuremethodsource{Pelican-Sim}{\cite{zou2026pelicansim}~{\normalfont\scriptsize\mbox{arXiv'26}}}
\measuremethodsource{DW05}{\cite{dexmal2026dw05}~{\normalfont\scriptsize\mbox{GitHub'26}}}
\measuremethodsource{BWM}{\cite{bwmteam2026bwm}~{\normalfont\scriptsize\mbox{arXiv'26}}}
\begin{resultstable}{@{}lcrrrrr@{}}
\toprule
Method & \shortstack{Extra\\input} & \shortstack{PSNR\\(dB)$\uparrow$} & \shortstack{SSIM\\$\uparrow$} & \shortstack{LPIPS\\$\downarrow$} & \shortstack{FID\\$\downarrow$} & \shortstack{FVD\\$\downarrow$} \\
\midrule
\multicolumn{7}{l}{\emph{Camera conditioning}} \\
\methodsource{RealCam-I2V}{\cite{li2025realcam}~{\normalfont\scriptsize\mbox{ICCV'25}}} & C & 16.34 & 0.535 & 0.462 & 96.04 & 841.3 \\
\methodsource{MotionCtrl}{\cite{wang2023motionctrl}~{\normalfont\scriptsize\mbox{SIGGRAPH'24}}} & C & 18.51 & 0.802 & 0.280 & 72.00 & 348.8 \\
\methodsource{CameraCtrl}{\cite{he2024cameractrl}~{\normalfont\scriptsize\mbox{ICLR'25}}} & C & 20.73 & 0.838 & 0.168 & 29.23 & 289.7 \\
\methodsource{TriMotion}{\cite{shin2026trimotion}~{\normalfont\scriptsize\mbox{ECCV'26}}} & C & 19.87 & 0.821 & 0.205 & 43.65 & 223.7 \\
\methodsource{DualCamCtrl}{\cite{zhang2025dualcamctrl}~{\normalfont\scriptsize\mbox{ECCV'26}}} & C & 20.92 & 0.853 & 0.161 & 26.54 & 147.7 \\
\methodsource{PPMem}{\cite{he2026ppmem}~{\normalfont\scriptsize\mbox{NeurIPS'26}}} & C & 21.41 & 0.864 & 0.144 & 21.48 & 130.5 \\
\midrule
\method  & C & 22.12 & 0.884 & 0.122 & 18.90 & 110.9 \\
\midrule
\multicolumn{7}{l}{\emph{Action conditioning}} \\
\methodsource{IRASim}{\cite{zhu2024irasim}~{\normalfont\scriptsize\mbox{ICCV'25}}} & A & 26.51 & 0.887 & 0.080 & 13.21 & 75.6 \\
\methodsource{Ctrl-World}{\cite{guo2026ctrl}~{\normalfont\scriptsize\mbox{ICLR'26}}} & A & 26.93 & 0.906 & 0.073 & 12.08 & 68.9 \\
\methodsource{Pelican-Sim}{\cite{zou2026pelicansim}~{\normalfont\scriptsize\mbox{arXiv'26}}} & C,A & 26.84 & 0.901 & 0.076 & 12.64 & 71.3 \\
\methodsource{DW05}{\cite{dexmal2026dw05}~{\normalfont\scriptsize\mbox{GitHub'26}}} & A & 27.04 & 0.910 & 0.071 & 11.83 & \underline{63.4} \\
\methodsource{BWM}{\cite{bwmteam2026bwm}~{\normalfont\scriptsize\mbox{arXiv'26}}} & A & \underline{27.20} & \underline{0.918} & \underline{0.068} & \textbf{11.25} & 64.8 \\
\midrule
\method  & C,A & \textbf{27.91} & \textbf{0.930} & \textbf{0.058} & \underline{11.53} & \textbf{55.1} \\
\bottomrule
\end{resultstable}
\end{table}

\begin{figure*}[!t]
    \centering
    \includegraphics[width=0.94\textwidth]{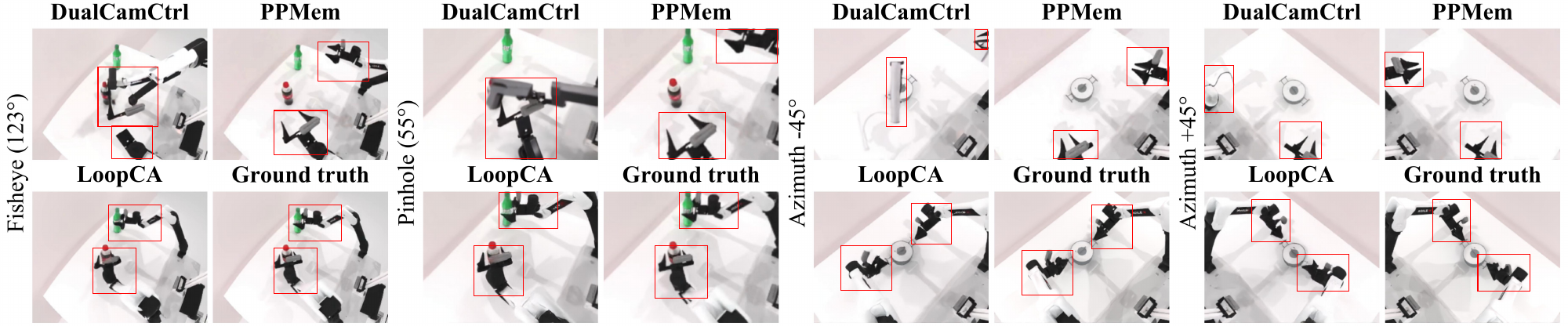}
    \caption{\textbf{Camera-control comparison.}
    Final frame under fisheye, pinhole, and rotated pinhole viewpoints.}
    \label{fig:camera_control_qualitative}
\end{figure*}

\paragraph{RGB prediction.}
As shown in Table~\ref{tab:external_comparison}, \method\ reaches
27.91\,dB PSNR and 0.930 SSIM, 0.71\,dB and 1.3\% higher than BWM, and
reduces LPIPS by 14.7\% versus BWM and FVD by 13.1\% versus DW05. The
improvement stems from rendering the supplied trajectory into RAAFs aligned
with the camera rays, so the action and camera guidance stay consistent. Without action conditioning,
\method\ still reaches 22.12\,dB PSNR and 0.884 SSIM, 0.71\,dB and 2.3\%
higher than PPMem.

\begin{table}[!t]
\centering
\footnotesize
\setlength{\tabcolsep}{3pt}
\caption{\textbf{Geometry prediction on Test Set.} The top group directly generates 3D representations, while the bottom group reconstructs them from the ground-truth video.}
\label{tab:pointmap_comparison}
\setlength{\methodsourcewidth}{0pt}
\measuremethodsource{DualCamCtrl}{\cite{zhang2025dualcamctrl}~{\normalfont\scriptsize\mbox{ECCV'26}}}
\measuremethodsource{PPMem}{\cite{he2026ppmem}~{\normalfont\scriptsize\mbox{NeurIPS'26}}}
\measuremethodsource{4DNeX}{\cite{chen20254dnex}~{\normalfont\scriptsize\mbox{arXiv'25}}}
\measuremethodsource{Geo4D}{\cite{jiang2025geo4d}~{\normalfont\scriptsize\mbox{ICCV'25}}}
\measuremethodsource{V-DPM}{\cite{sucar2026vdpm}~{\normalfont\scriptsize\mbox{CVPR'26}}}
\measuremethodsource{$\pi^3$}{\cite{wang2025pi}~{\normalfont\scriptsize\mbox{ICLR'26}}}
\measuremethodsource{CUT3R}{\cite{wang2025cut3r}~{\normalfont\scriptsize\mbox{CVPR'25}}}
\begin{resultstable}{lrr}
\toprule
Output / RGB & \shortstack{AbsErr\\(cm)$\downarrow$} & \shortstack{$\delta_p$\\(\%)$\uparrow$} \\
\midrule
\multicolumn{3}{l}{\emph{Direct 3D generation}} \\
\methodsource{4DNeX}{\cite{chen20254dnex}~{\normalfont\scriptsize\mbox{arXiv'25}}} & 14.94 & 79.0 \\
\methodsource{DualCamCtrl}{\cite{zhang2025dualcamctrl}~{\normalfont\scriptsize\mbox{ECCV'26}}} & 12.83 & 83.2 \\
\methodsource{PPMem}{\cite{he2026ppmem}~{\normalfont\scriptsize\mbox{NeurIPS'26}}} & 6.80 & 81.3 \\

\midrule
\multicolumn{3}{l}{\emph{GT-video reconstruction}} \\
\methodsource{Geo4D}{\cite{jiang2025geo4d}~{\normalfont\scriptsize\mbox{ICCV'25}}} & 8.45 & 55.3 \\
\methodsource{CUT3R}{\cite{wang2025cut3r}~{\normalfont\scriptsize\mbox{CVPR'25}}} & 7.71 & 64.9 \\
\methodsource{$\pi^3$}{\cite{wang2025pi}~{\normalfont\scriptsize\mbox{ICLR'26}}} & 7.47 & 71.7 \\
\methodsource{V-DPM}{\cite{sucar2026vdpm}~{\normalfont\scriptsize\mbox{CVPR'26}}} & \underline{6.32} & \underline{84.3} \\
\midrule
\method\  & \textbf{3.67} & \textbf{90.1} \\
\bottomrule
\end{resultstable}
\end{table}

\begin{figure*}[!t]
    \centering
    \includegraphics[width=0.94\textwidth]{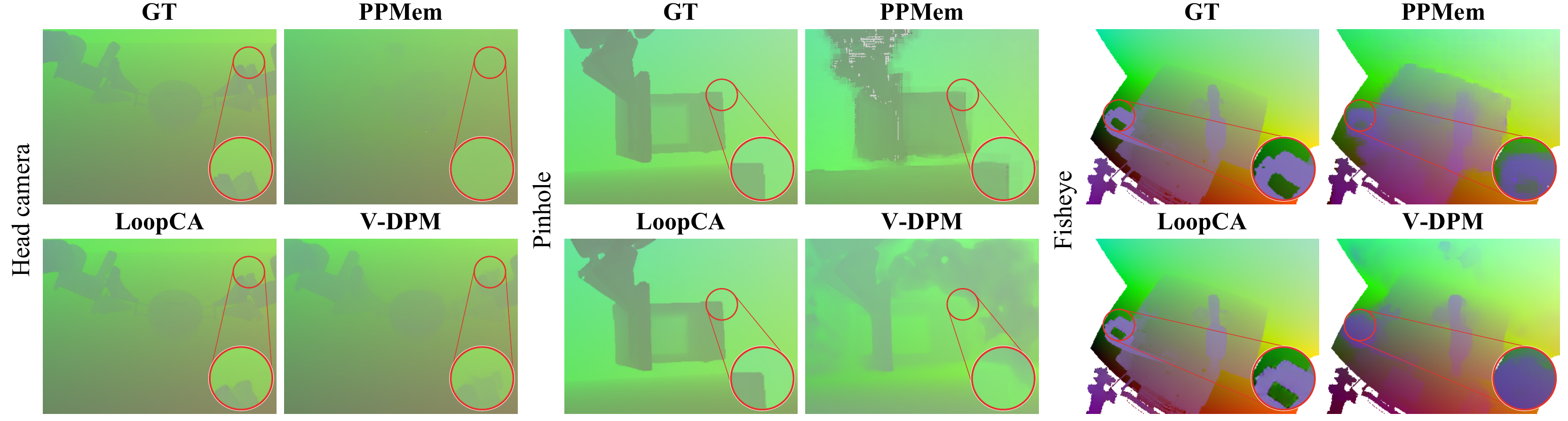}
    \caption{\textbf{Pointmap comparison.}
    Final frame; RGB channels encode world-coordinate $X,Y,Z$.}
    \label{fig:pointmap_qualitative}
\end{figure*}

\paragraph{Geometry prediction.}
Table~\ref{tab:pointmap_comparison} reports geometry results, where
\method\ reaches 3.67\,cm point error and 90.1\% $\delta_p$, 46.0\% lower
and 8.8 percentage points higher than PPMem. Reconstructing from the
observed ground-truth video, V-DPM reaches 6.32\,cm, so our native pointmap
is 41.9\% lower. This shows \method\ has strong geometry prediction
capability. The lower error follows from predicting pointmaps natively in
the shared latent space and coupling the RGB stream to geometry through
laterals, rather than reconstructing geometry from generated RGB post hoc.

\begin{table}[!t]
\centering
\footnotesize
\setlength{\tabcolsep}{1.8pt}
\caption{\textbf{Camera control on Pose Set.}}
\label{tab:camera_pose}
\begin{resultstable}{@{}lrrrr@{}}
\toprule
Method & \shortstack{RotErr\\($^\circ$)$\downarrow$} & \shortstack{TransErr\\(cm)$\downarrow$} & FID $\downarrow$ & FVD $\downarrow$ \\
\midrule
MotionCtrl~\cite{wang2023motionctrl} & 8.22 & 3.69 & 52.09 & 368.2 \\
RealCam-I2V~\cite{li2025realcam} & 2.59 & 2.15 & 83.00 & 745.7 \\
Pelican-Sim~\cite{zou2026pelicansim} & 5.04 & 2.77 & 63.16 & 470.7 \\
TriMotion~\cite{shin2026trimotion} & 4.52 & 2.19 & 42.12 & 288.7 \\
CameraCtrl~\cite{he2024cameractrl} & 3.73 & 2.62 & 34.58 & 390.1 \\
PPMem~\cite{he2026ppmem} & 3.64 & \underline{1.71} & \underline{30.12} & 233.7 \\
DualCamCtrl~\cite{zhang2025dualcamctrl} & \underline{2.55} & 1.82 & \underline{30.12} & \underline{225.4} \\
\midrule
\method{} & \textbf{2.17} & \textbf{1.45} & \textbf{25.60} & \textbf{191.6} \\
\bottomrule
\end{resultstable}
\end{table}

\paragraph{Camera control.}
As Table~\ref{tab:camera_pose} shows, on the Pose Set \method\ reaches
2.17$^\circ$ rotation error and 1.45\,cm translation error, 14.9\% and
15.2\% lower than DualCamCtrl and PPMem, respectively. Across the 123
moving-camera clips, FID is 15.0\% lower than PPMem and DualCamCtrl (tied),
and FVD is 15.0\% lower than DualCamCtrl. The unified Pl\"ucker ray field, which
encodes intrinsics and extrinsics for both pinhole and fisheye cameras,
drives this improvement.

\begin{figure*}[!t]
    \centering
    \includegraphics[width=0.94\textwidth]{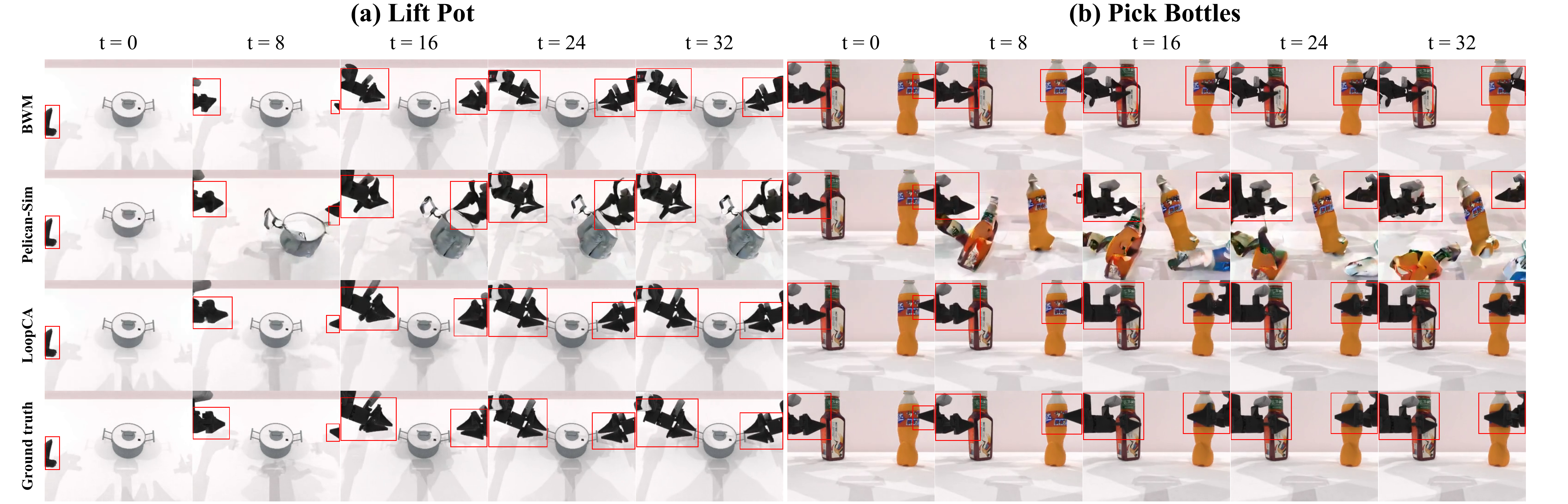}
    \caption{\textbf{Action-conditioned comparison.}
    }
    \label{fig:action_control_qualitative}
\end{figure*}

\begin{table}[!t]
\centering
\footnotesize
\setlength{\tabcolsep}{4pt}
\caption{\textbf{Action control on Test Set.}}
\label{tab:action_control}
\begin{resultstable}{@{}lrrr@{}}
\toprule
Method & TrajAcc $\uparrow$ & EEFErr (px) $\downarrow$ & PCK-AUC (\%) $\uparrow$ \\
\midrule
DW05~\cite{dexmal2026dw05} & 0.3065 & 9.29 & 86.09 \\
Pelican-Sim~\cite{zou2026pelicansim} & 0.2927 & 9.95 & 85.24 \\
BWM~\cite{bwmteam2026bwm} & \underline{0.3180} & \underline{8.90} & \underline{86.84} \\
\midrule
\method{} & \textbf{0.3745} & \textbf{7.59} & \textbf{87.98} \\
\bottomrule
\end{resultstable}
\end{table}

\paragraph{Action control.}
Table~\ref{tab:action_control} summarizes action control, where \method\
reaches 7.59\,px EEFErr, 87.98\% PCK-AUC, and 0.3745 TrajAcc, 14.7\%
lower, 1.14 percentage points higher, and 17.8\% higher than BWM,
respectively. Rendering the trajectory into image space as RAAFs underlies
this gain, since the model conditions on motion that is aligned with the
generated image.

\begin{table}[!t]
\centering
\footnotesize
\setlength{\tabcolsep}{2.5pt}
\caption{\textbf{Efficiency on Test Set.} $L$ is the number of loops.}
\label{tab:efficiency_comparison}
\label{tab:loop_depth}
\begin{resultstable}{@{}lcccc@{}}
\toprule
Method & Params (B) & Compute (TFLOPs) & Time (s) & PSNR (dB)$\uparrow$ \\
\midrule
PPMem~\cite{he2026ppmem} & 5.586 & 27293.00 & 128.85 & 21.41 \\
BWM~\cite{bwmteam2026bwm} & 5.026 & 4106.77 & 18.42 & 27.20 \\
Non-looped baseline & 2.148 & 1129.58 & 7.34 & 27.91 \\
\method ($L=1$) & 1.381 & 664.19 & 4.40 & 27.05 \\
\method ($L=2$) & 1.381 & 819.32 & 5.32 & 27.76 \\
\method ($L=3$) & 1.381 & 974.45 & 6.28 & 27.90 \\
\method ($L=4$) & 1.381 & 1129.58 & 7.16 & 27.91 \\
\bottomrule
\end{resultstable}
\end{table}

\paragraph{Efficiency.}
As shown in Table~\ref{tab:efficiency_comparison}, \method\ uses 72.5\% fewer
parameters than BWM. Compared with a non-looped baseline of the same
compute, the looped model uses 35.7\% fewer parameters and reaches the same PSNR. Two-loop
inference cuts TFLOPs by 27.5\% and time by 25.7\% versus four loops with
only a 0.15\,dB PSNR drop.
\method reduces parameters and supports adjustable inference depth for
lower latency. These benefits come from reusing blocks across streams and
depth with logical-depth-specific conditioning and multi-exit supervision.

\begin{table*}[!t]
\centering
\footnotesize
\setlength{\tabcolsep}{3pt}
\caption{\textbf{Quality and cost versus loop count on Test Set.} }
\label{tab:number_of_loops}
\begin{resultstable}{@{}lrrrrrrrrrrrr@{}}
\toprule
Number of loops $L$ & 1 & 2 & 3 & 4 & 5 & 6 & 8 & 10 & 12 & 15 & 17 & 20 \\
\midrule
PSNR (dB)$\uparrow$ & 27.05 & 27.76 & 27.90 & 27.91 & 27.40 & 26.75 & 25.48 & 24.23 & 23.03 & 21.10 & 19.61 & 17.23 \\
AbsErr (cm)$\downarrow$ & 4.04 & 3.79 & 3.69 & 3.67 & 3.82 & 4.24 & 5.22 & 6.32 & 7.48 & 9.49 & 11.09 & 14.41 \\
Compute (TFLOPs)$\downarrow$ & 664.19 & 819.32 & 974.45 & 1129.58 & 1284.72 & 1439.85 & 1750.11 & 2060.38 & 2370.64 & 2836.04 & 3146.31 & 3611.70 \\
Time (s)$\downarrow$ & 4.40 & 5.32 & 6.28 & 7.16 & 8.08 & 8.99 & 10.86 & 12.69 & 14.55 & 17.32 & 19.18 & 21.96 \\
\bottomrule
\end{resultstable}
\end{table*}

\begin{table*}[!t]
\begin{minipage}[t]{\dimexpr\textwidth/2-\columnsep/2\relax}
\centering
\footnotesize
\setlength{\tabcolsep}{2.5pt}
\caption{\textbf{Ablation study on Test Set.} $L$ is the number of loops.}
\label{tab:raaf_input_ablation}
\label{tab:component_ablation}
\begin{resultstable}{@{}p{.38\columnwidth}rrrr@{}}
\toprule
Variant & \shortstack{PSNR\\(dB)$\uparrow$} & FID $\downarrow$ & FVD $\downarrow$ & \shortstack{AbsErr\\(cm)$\downarrow$} \\
\midrule
 non-looped, 30 blocks & 27.91 & 11.53 & 55.2 & 3.67 \\
 non-looped, 15 blocks & 22.74 & 24.25 & 125.7 & 8.80 \\
 \midrule
w/o RAAF  & 22.12 & 18.90 & 110.9 & 5.78 \\
\midrule
Loss: final-only, $L=2$  & 21.52 & 40.34 & 197.5 & 14.67 \\
Loss: final-only, $L=4$  & 27.91 & 11.52 & 55.0 & 3.67 \\
Loss: multi-exit, $L=2$ & 27.76 & 12.50 & 59.2 & 3.79 \\
\midrule
w/o lateral  & 27.91 & 11.53 & 55.1 & 102.45 \\
\midrule
\method & 27.91 & 11.53 & 55.1 & 3.67 \\
\bottomrule
\end{resultstable}

\end{minipage}\hfill
\begin{minipage}[t]{\dimexpr\textwidth/2-\columnsep/2\relax}
\centering
\footnotesize
\setlength{\tabcolsep}{2.5pt}
\caption{\textbf{Downstream $\pi_{0.5}$ success rates (\%).} +GT/+LC use ground-truth/\method-generated views; $\Delta$ is the mean gain over original-only training.}

\label{tab:downstream_vla}
\begin{resultstable}{@{}lcrrrrrr@{}}
\toprule
Data & $\theta$ & Scale & \shortstack{Pill\\Bottle} & Bread & \shortstack{Burger/\\Fries} & Mean & \shortstack{$\Delta$\\(pp)} \\
\midrule
Orig. & $0^\circ$ & 85.0 & 90.0 & 55.0 & 100.0 & 82.5 &  \\
 & $\pm10^\circ$ & 25.0 & 20.0 & 10.0 & 20.0 & 18.8 &  \\
\midrule
+GT & $0^\circ$ & 80.0 & 90.0 & 80.0 & 100.0 & 87.5 & +5.0 \\
 & $\pm10^\circ$ & 82.5 & 87.5 & 55.0 & 97.5 & 80.6 & +61.8 \\
\midrule
+LC & $0^\circ$ & 75.0 & 95.0 & 80.0 & 95.0 & 86.3 & +3.8 \\
 & $\pm10^\circ$ & 80.0 & 82.5 & 65.0 & 95.0 & 80.6 & +61.8 \\
\bottomrule
\end{resultstable}

\end{minipage}
\end{table*}

\subsection{Qualitative Results}
\label{sec:qualitative_results}

We report qualitative results for camera control, action control, and
pointmap prediction. Figure~\ref{fig:camera_control_qualitative} shows that
\method\ renders RGB frames that remain coherent with the reference scene
 under fisheye-pinhole lens conversion and
rotated viewpoints, and Figure~\ref{fig:pointmap_qualitative} shows pointmaps
whose world-coordinate geometry aligns with the visible surfaces across
camera families. Figure~\ref{fig:action_control_qualitative} shows that
action-conditioned predictions keep the robot motion consistent with the trajectory over time. These results demonstrate that
\method\  synthesizes consistent appearance and geometry under
generalized camera-action control.

\subsection{Ablation Studies}
\label{sec:ablations}
\label{sec:analysis}
\label{sec:loop_results}
\label{sec:e1_control}

\paragraph{Number of loops.}
Table~\ref{tab:number_of_loops} sweeps the loop count, where four loops
give the best result, and reducing inference from four to two cuts latency
by 25.7\% with 0.15\,dB lower PSNR and 1.19\,mm more point error.

\paragraph{Architecture and training.}
Table~\ref{tab:component_ablation} ablates the backbone, conditioning, and
coupling. A non-looped 15-block
backbone drops PSNR by 5.17\,dB and raises AbsErr to 8.80\,cm
versus the 30-block backbone, while reusing the shared core four times
recovers this depth and reaches 27.91\,dB and 3.67\,cm. Omitting RAAF
lowers PSNR by 5.79\,dB and raises AbsErr from 3.67 to 5.78\,cm, showing
the model relies on the rendered action fields. Multi-exit supervision is
essential for shallow exits: a final-only $L{=}2$ exit reaches only
21.52\,dB and 14.67\,cm, whereas the multi-exit $L{=}2$ exit reaches
27.76\,dB and 3.79\,cm. Disabling the RGB-to-pointmap lateral at inference
leaves RGB quality unchanged but collapses geometry to 102.45\,cm AbsErr,
showing the geometry stream relies on the trained cross-modal coupling
rather than on the RGB stream alone.

\subsection{Downstream Policy Learning}
\label{sec:downstream_vla}

Table~\ref{tab:downstream_vla} summarizes the downstream policy results. We
finetune $\pi_{0.5}$~\cite{black2025pi05} with LoRA~\cite{hu2021lora} across four representative manipulation tasks using
expert data augmented by \method-generated camera views, comparing against
original-only training and GT-view augmentation. Under $\pm10^\circ$ head-camera shifts, generated-view
augmentation raises mean success from 18.8\% to 80.6\%, a gain of 61.8
percentage points over original-only training. This gain comes from camera-diverse generated views
paired with consistent action labels, demonstrating that \method\ supports
downstream policy learning.

\FloatBarrier
\section{Conclusion}
\label{sec:conclusion}

We presented \method, a looped dual-stream diffusion Transformer for efficient
and accurate joint 4D world modeling under generalized camera-action control.
\method\ couples RGB and pointmap generation in a shared latent space with
geometry adapters and stop-gradient laterals.
It reuses a looped core with logical-depth-specific conditioning and
multi-exit supervision, reducing model parameters by 35.7\% relative to
the non-looped baseline and supporting adjustable inference depth.
Camera and action signals are aligned through per-pixel Pl\"ucker ray
fields and rendered RAAFs.
On a camera-diverse RoboTwin benchmark, \method\ achieves 27.91\,dB PSNR and
3.67\,cm 3D point error, reduces latency by 25.7\% when selecting two instead of four loops, at
only 0.15\,dB PSNR cost, and boosts downstream manipulation success by 61.8
percentage points. These results establish \method\ as a parameter-efficient
and camera-action-controllable 4D world model. Future work could apply the rendering-aligned camera-action control to other
world-model backbones and explore the use of looped computation for historical
frames and additional sensors.

\section*{LLM Usage Statement}
We used OpenAI's ChatGPT and Anthropic's Claude to assist
with editing and refining author-written text, including
related work based on author-selected references. The tools
also assisted with reviewing, debugging, and refining
author-developed code for training, evaluation, and
efficiency measurement, as well as experimental analysis
and figure preparation. No LLM contributed research ideas,
hypotheses, claims, or conclusions, and none was used for
data creation or annotation. The manuscript was not
submitted to an LLM for reviewer-style feedback.
All LLM-assisted content was verified by the authors,
who take full responsibility for the submission.

{
    \small
    \bibliographystyle{ieeenat_fullname}
    \bibliography{main}
}

\end{document}